\documentclass{article}

\usepackage{microtype}
\usepackage{graphicx}
\usepackage{subcaption}
\usepackage{booktabs} 
\usepackage{xspace}
\usepackage{makecell}

\usepackage{hyperref}

\usepackage[accepted]{icml2018}

\icmltitlerunning{An Analysis of Neural Language Modeling at Multiple Scales}

\newcommand{\enwik}{\texttt{enwik8}\xspace}

\begin{document}

\twocolumn[
\icmltitle{An Analysis of Neural Language Modeling at Multiple Scales}



\icmlsetsymbol{equal}{*}

\begin{icmlauthorlist}
\icmlauthor{Stephen Merity}{to,equal}
\icmlauthor{Nitish Shirish Keskar}{to,equal}
\icmlauthor{Richard Socher}{to}
\end{icmlauthorlist}

\icmlaffiliation{to}{Salesforce Research, Palo Alto, CA -- 94301}
\icmlcorrespondingauthor{Stephen Merity}{smerity@salesforce.com}
\icmlkeywords{language modeling, LSTM, QRNN, RNN}

\vskip 0.3in
]



\printAffiliationsAndNotice{\icmlEqualContribution} 

\begin{abstract}
Many of the leading approaches in language modeling introduce novel, complex and specialized architectures.
We take existing state-of-the-art word level language models based on LSTMs and QRNNs and extend them to both larger vocabularies as well as character-level granularity.
When properly tuned, LSTMs and QRNNs achieve state-of-the-art results on character-level (Penn Treebank, \enwik) and word-level (WikiText-103) datasets, respectively.
Results are obtained in only 12 hours (WikiText-103) to 2 days (\enwik) using a single modern GPU.
\end{abstract}

\section{Introduction}

Language modeling (LM) is one of the foundational tasks of natural language processing.
The task involves predicting the $(n+1)^{th}$ token in a sequence given the $n$ preceding tokens.
Trained LMs are useful in many applications including speech recognition \citep{yu2014automatic}, machine translation \citep{koehn2009statistical}, natural language generation \citep{radford2017learning, smerity-revisiting}, learning token embeddings, and as a general-purpose feature extractor for downstream tasks \cite{elmo}.

Language models can operate at various granularities, with these tokens formed from either words, sub-words, or characters.
While the underlying objective remains the same across all sub-tasks, each has its own unique set of benefits and challenges.

In practice, word-level LMs appear to perform better at downstream tasks compared to character-level LMs but suffer from increased computational cost due to large vocabulary sizes.
Even with a large vocabulary, word-level LMs still need to replace infrequent words with out-of-vocabulary (OoV) tokens.
Character-level LMs do not suffer this OoV problem as the potentially infinite set of potential words can be constructed by repeatedly selecting from a limited set of characters.
This introduces two issues for character-level LMs however - they are slower to process than their word-level counterparts as the number of tokens increases substantially, and due to this, they experience more extreme issues with vanishing gradients.
Tokens are also far less informative in a character-level LM.
For a word-level LM to understand the likely topics a sentence may cover, one or two discriminative words may be sufficient.
In comparison, a character-level LM would require at least half a dozen.
Given this distinction, word- and character-level LMs are commonly trained with vastly different architectures.
While vanilla Long Short Term Memory (LSTM) networks have been shown to achieve state-of-the-art performance on word-level LM they have been hitherto considered insufficient for competitive character-level LM; see e.g., \cite{Melis2017}
In this paper, we show that a given baseline model framework, composed of a vanilla RNN (LSTM or its cheaper counterpart, QRNN \cite{Bradbury2016}) and an adaptive softmax, is capable of modeling both character- and word-level tasks at multiple scales of data whilst achieving state-of-the-art results.
Further, we present additional analysis regarding a comparison of the LSTM and QRNN architectures and the importance of various hyperparameters in the model. We conclude the paper with a discussion concerning the choice of datasets and model metrics. 

\section{Motivation}

Recent research has shown that a well tuned LSTM baseline can outperform more complex architectures in the task of word-level language modeling \citep{merity2018regularizing,Melis2017}. 
The model in \citet{merity2018regularizing} also aims to use well optimized components, such as the NVIDIA cuDNN LSTM or highly parallel Quasi-Recurrent Neural Network \citep{Bradbury2016}, allowing for rapid convergence and experimentation due to efficient hardware usage.
While the models were shown to achieve state-of-the-art results on modest language modeling datasets, their application to larger-scale word-level language modeling or character-level language modeling had not been successful.

Large-scale word- and character-level datasets can both require training over hundreds of millions of tokens.
This requires an efficient model such that experimentation does not require vast amounts of time or resources.
From this, we aim to ensure our model can train in a matter of days or hours on a single modern GPU and can achieve results competitive with current state-of-the-art results.

\section{Model architecture}

Our underlying architecture is based upon the model used in \citet{merity2018regularizing}.
Their model consists of a trainable embedding layer, one or more layers of a stacked recurrent neural network, and a softmax classifier.
The embedding and softmax classifier utilize tied weights \cite{Inan2016,Press2016} to both decrease the total parameter count as well as improve classification accuracy over rare words.
Their experimental setup features various optimization and regularization variants such as randomized-length backpropagation through time (BPTT),
embedding dropout, variational dropout, activation regularization (AR), and
temporal activation regularization (TAR).
This model framework has been shown to achieve state-of-the-art results using either an LSTM or QRNN based model in under a day on an NVIDIA Quadro GP100.

\subsection{LSTM and QRNN}

In \citet{merity2018regularizing}, two different recurrent neural network cells are evaluated: the Long Short Term Memory (LSTM) \citep{Hochreiter1997LongSM} and the Quasi-Recurrent Neural Network (QRNN) \citep{Bradbury2016}.

While the LSTM works well in terms of task performance, it is not an optimal fit for ensuring high GPU utilization.
The LSTM is sequential in nature, relying on the output of the previous timestep before work can begin on the current timestep.
This limits the concurrency of the LSTM to the size of the batch and even then can result in substantial CUDA kernel overhead if each timestep must be processed individually.

The QRNN attempts to improve GPU utilization for recurrent neural networks in two ways: it uses convolutional layers for processing the input, which apply in parallel across timesteps, and then uses a minimalist recurrent pooling function that applies in parallel across channels.
As the convolutional layer does not rely on the output of the previous timestep, all input processing can be batched into a single matrix multiplication.
While the recurrent pooling function is sequential, it is a fast element-wise operation that is applied to existing prepared inputs, thus producing next to no overhead.
In our investigations, the overhead of applying dropout to the inputs was more significant than the recurrent pooling function.

The QRNN can be up to 16 times faster than the optimized NVIDIA cuDNN LSTM for timings only over the RNN itself \citep{Bradbury2016}.
When two networks are composed of approximately equal size with the LSTM and QRNN, \cite{merity2018regularizing} found that QRNN based models were overall $2-4\times$ faster per epoch.
For word-level LMs, QRNNs were also found to require fewer epochs to converge and achieved comparable state-of-the-art results on the word-level Penn Treebank and WikiText-2 datasets.

Given the sizes of the datasets we intend to process can be many millions of tokens in length, the potential speed benefit of the QRNN is of interest, especially if the results remain competitive to that of the LSTM.

\subsection{Longer BPTT lengths}

Truncated backpropagation through time (BPTT) \citep{williams1990efficient,werbos1990backpropagation} is necessary for long form continuous datasets.
Traditional LMs use relatively small BPTT windows of 50 or less for word-level LM and 100 or less for character-level.
For both granularities, however, the gradient approximation made by truncated BPTT is highly problematic.
A single sentence may well be longer than the truncated BPTT window used in character- or even word-level modeling, preventing useful long term dependencies from being discovered.
This is exacerbated if the long term dependencies are split across paragraphs or pages.
The longer the sequence length used during truncated BPTT, the further back long term dependencies can be explicitly formed, potentially benefiting the model's accuracy.

In addition to a potential accuracy benefit, longer BPTT windows can improve GPU utilization.
In most models, the primary manner to improve GPU utilization is through increasing the number of examples per batch, shown to be highly effective for tasks such as image classification \citep{goyal2017accurate}.
Due to the more parallel nature of the QRNN, the QRNN can process long sequences entirely in parallel, in comparison to the more sequential LSTM.
This is as the QRNN does not feature a slow sequential hidden-to-hidden matrix multiplication at each timestep, instead relying on a fast element-wise operator.

\subsection{Tied adaptive softmax}

Large vocabulary sizes are a major issue on large-scale word-level datasets and can result in impractically slow models (softmax overhead) or models that are impossible to train due to a lack of GPU memory (parameter overhead).
To address these issues, we use a modified version of the adaptive softmax \citep{grave2016efficient}, extended to allow for tied weights \cite{Inan2016,Press2016}.
While other softmax approximation strategies exist \citep{morin,nce}, the adaptive softmax has been shown to achieve results close to that of the full softmax whilst maintaining high GPU efficiency.

The adaptive softmax uses a hierarchy determined by word frequency to reduce computation time.
Words are split into two levels: the first level (the short-list) and the second level, which forms clusters of rare words.
Each cluster has a representative token in the short-list which determines the overall probability assigned across all words within that cluster.
Due to Zipf's law, most words will only require a softmax over the short-list during training, reducing computation and memory usage.
Clusters on the second level are also constructed such that the matrix multiplications required for a standard batch are near optimal for GPU efficiency.

In the original adaptive softmax implementation, words within clusters on the second level feature a reduced embedding size, reducing from embedding dimensionality $d$ to $\frac{d}{4}$.
This minimizes both total parameter count and the size of the softmax matrix multiplications.
The paper justifies this by noting that rare words are unlikely to need full embedding size fidelity due to how infrequently they occur.

Weight tying cannot be used with this memory optimization however.
As weight tying re-uses the word vectors from the embedding as the targets in the softmax, the embedding dimensionality must be equal for all words.
We discard the adaptive softmax memory optimization in order to utilize tied weights.
Counter-intuitively, weight tying actually reduces the memory usage even more than the adaptive softmax memory optimization, as weight tying allows us to halve the memory usage by re-using the word vectors from the embedding layer.

When a dataset is processed using a small vocabulary, all the words can be placed into the short-list, thus reducing the adaptive softmax to a standard softmax.

\begin{table*}
\begin{center}
\begin{tabular}[t]{c|ccc|ccc|ccc}
\toprule
& \multicolumn{3}{c|}{{\bf Penn Treebank (Character)}} & \multicolumn{3}{c|}{{\bf \enwik}} & \multicolumn{3}{c}{{\bf WikiText-103}} \\
& Train & Valid & Test & Train & Valid & Test & Train & Valid & Test \\
\midrule
Tokens & 5.01M & 393k & 442k & 90M & 5M & 5M & 103.2M & 217k & 245k \\
\midrule
Vocab size & \multicolumn{3}{c|}{51} & \multicolumn{3}{c|}{205} & \multicolumn{3}{c}{267,735} \\
OoV rate & \multicolumn{3}{c|}{-} & \multicolumn{3}{c|}{-} & \multicolumn{3}{c}{0.4\%} \\
\bottomrule
\end{tabular}
\caption{Statistics of the character-level Penn Treebank, character-level \enwik dataset, and WikiText-103. The out of vocabulary (OoV) rate notes what percentage of tokens have been replaced by an $\langle unk \rangle$ token, not applicable to character-level datasets.}
\label{table:data}
\end{center}
\end{table*}

\section{Experiments}

Our work is over three datasets, two character-level datasets (Penn Treebank, \enwik) and a large-scale word-level (WikiText-103) dataset.
A set of overview statistics for these datasets are presented in Table \ref{table:data}.

\begin{table}
\begin{center}
 \begin{tabular}{@{}l r r@{}}
\toprule[1.5pt]
Model & BPC & Params \\
\midrule
 Zoneout LSTM \cite{Krueger2016}& 1.27 & -\\
 2-Layers LSTM \cite{fastslowlm} & 1.243 & 6.6M\\
 HM-LSTM \cite{chung2016hierarchical}& 1.24 & - \\
 HyperLSTM - small \cite{ha2016hypernetworks}& 1.233 & 5.1M \\
 HyperLSTM \cite{ha2016hypernetworks} & 1.219 & 14.4M \\
 NASCell (small) \cite{Zoph2016} & 1.228 & 6.6M \\
 NASCell \cite{Zoph2016} & 1.214 & 16.3M \\
 FS-LSTM-2 \cite{fastslowlm} & 1.190 & 7.2M \\
 FS-LSTM-4 \cite{fastslowlm} & 1.193 & 6.5M \\
 \midrule
 6 layer QRNN ($h=1024$) (ours) & 1.187 & 13.8M \\
 3 layer LSTM ($h=1000$) (ours) & 1.175 & 13.8M \\
 \bottomrule
\end{tabular}
\end{center}
\caption{Bits Per Character (BPC) on character-level Penn Treebank.}
\label{table:ptb}
\end{table}

\begin{table}
\begin{center}
 \begin{tabular}{@{} llr @{}}
 \toprule[1.5pt]
 Model & BPC & Params \\
 \midrule
 LSTM, $2000$ units \cite{fastslowlm} & 1.461 & 18M \\
 Layer Norm LSTM, $1800$ units & 1.402 & 14M\\
 HyperLSTM \cite{ha2016hypernetworks} & 1.340 & 27M \\
 HM-LSTM \cite{chung2016hierarchical}& 1.32 & 35M \\
 SD Zoneout \cite{rocki2016surprisal} & 1.31 & 64M \\
 RHN - depth 5 \cite{Zilly2016}& 1.31 & 23M \\
 RHN - depth 10  \cite{Zilly2016}& 1.30 & 21M \\
 Large RHN \cite{Zilly2016}& 1.27 & 46M \\
 FS-LSTM-2 \cite{fastslowlm} & 1.290 & 27M \\
 FS-LSTM-4 \cite{fastslowlm} & 1.277 & 27M \\
 Large FS-LSTM-4 \cite{fastslowlm} & 1.245 & 47M \\
 \midrule
 4 layer QRNN ($h=1800$) (ours) & 1.336 & 26M \\
 3 layer LSTM ($h=1840$) (ours) & 1.232 & 47M \\
 \midrule
 \texttt{cmix} v13 \cite{cmix}  & 1.225 & - \\
 \bottomrule
\end{tabular}
\end{center}
\caption{Bits Per Character (BPC) on \enwik.}
\label{table:enwik8}
\end{table}

\subsection{Penn Treebank}

In \citet{Mikolov2010}, the Penn Treebank dataset \citep{marcus1994penn} was processed to generate a character-level language modeling dataset.
The dataset exists in both a word- and character-level form.

While the dataset was originally composed of Wall Street Journal articles, the preprocessed dataset removes many features considered important for capturing language modeling.
Oddly, the vocabulary of the words in the character-level dataset is limited to 10,000 - the same vocabulary as used in the word level dataset.
This vastly simplifies the task of character-level language modeling as character transitions will be limited to those found within the limited word level vocabulary.

In addition to the limited vocabulary, the character-level dataset is all lower case, has all punctuation removed (other than as part of certain words such as \verb+u.s.+ or \verb+mr.+), and replaces all numbers with \verb+N+.
All of these would be considered important subtasks for a character-level language model to cover.

When rare words are encountered in both the character and word level datasets, they're replaced with the token \verb+<unk>+.
This makes little sense for the character-level dataset, which doesn't suffer from out of vocabulary issues as the word level dataset would.
As the \verb+<unk>+ token is the only token to use the \verb+<+ and \verb+>+ characters, a sufficiently advanced model should always output \verb+unk>+ upon seeing \verb+<+ as this is the only time angle brackets are used.

We later explore these issues and their impact by comparing models trained on character-level Penn Treebank with a more realistic character-level dataset, \enwik.

We report our results, using the bits per character (BPC) metric, in Table~\ref{table:ptb}.
The model uses a three-layered LSTM with a BPTT length of $150$, embedding sizes of $200$ and hidden layers of size $1000$.
We regularize the model using LSTM dropout, weight dropout \cite{merity2018regularizing}, and weight decay.
These values are tuned coarsely.
For full hyper-parameter values, refer to Table~\ref{table:fullhypers}.
We train the model using the Adam \cite{kingma2014adam} optimizer with a learning rate of $2.7 \times 10^{-3}$ for $500$ epochs and reduce the learning rate by $10$ on epochs $300$ and $400$.
While the model was not optimized for convergence speed, it takes only 8 hours to train on an NVIDIA Volta GPU.
Both our LSTM and QRNN model beat the current state-of-the-art though using more parameters.
Of note is that the QRNN model uses 6 layers to achieve the same result as the LSTM with only 3 layers, suggesting that the limited recurrent computation capacity of the QRNN may be an issue.

\subsection{Hutter Wikipedia Prize (\enwik)}

The Hutter Prize Wikipedia dataset \cite{hutter}, also known as \enwik, is a byte-level dataset consisting of the first 100 million bytes of a Wikipedia XML dump.
For simplicity we shall refer to it as a character-level dataset.
Within these 100 million bytes are 205 unique tokens.
The Hutter Prize was launched in 2006 and is focused around compressing the \enwik dataset as efficiently as possible.

The XML dump contains a wide array of content, including English articles, XML data, hyperlinks, and special characters.
As the data has not been processed, it features many of the intricacies of language modeling, both natural and artificial, that we would like our models to capture.
For our experiments, we follow a standard setup where the train, validation and test sets consist of the first 90M, 5M, and 5M characters, respectively.

We report our results in Table~\ref{table:enwik8}.
The model uses a three-layered LSTM with a BPTT length of $200$, embedding sizes of $400$ and hidden layers of size $1850$.
The only explicit regularizer we employ is the weight dropped LSTM \cite{merity2018regularizing} of magnitude $0.2$.
For full hyper-parameter values, refer to Table~\ref{table:fullhypers}.
We train the model using the Adam \cite{kingma2014adam} optimizer with default hyperparameter for $50$ epochs and reduce the learning rate by $10$ on epochs $25$ and $35$.

This dataset is far more challenging than character-level Penn Treebank for multiple reasons.
The dataset has 18 times more training tokens and the data has not been processed at all, maintaining far more complex character to character transitions than that of the character-level Penn Treebank.
Potentially as a result of this, the QRNN based model underperforms models with comparable numbers of parameters.
The limited recurrent computation capacity of the QRNN appears to become a major issue when moving toward realistic character-level datasets.

\subsection{WikiText}
The WikiText-2 (WT2) and WikiText-103 (WT103) datasets introduced in \citet{Merity2016} contain lightly preprocessed Wikipedia articles, retaining the majority of punctuation and numbers.
The WT2 data set contains approximately 2 million words in the training set and 0.2 million in validation and test sets.
The WT103 data set contains a larger training set of 103 million words and the same validation and testing set as WT2.
As the Wikipedia articles are relatively long and are focused on a single topic, capturing and utilizing long term dependencies can be key to models obtaining strong performance. 

The underlying model used in this paper achieved state-of-the-art on language modeling for the word-level PTB and WikiText-2 datasets \cite{merity2018regularizing}
We show that, in conjunction with the tied adaptive softmax, we achieve state-of-the-art perplexity on the WikiText-103 data set using the AWD-QRNN framework; see Table~\ref{table:wikitext103}.
We opt to use the QRNN as the LSTM is 3 times slower, as reported in Table~\ref{table:wt103time}, and a QRNN based model's performance on word-level datasets has been found to equal that of an LSTM based model \citet{merity2018regularizing}.

By utilizing the QRNN and tied adaptive softmax, we were able to train to a state-of-the-art result with a NVIDIA Volta GPU in 12 hours.
The model used consisted of a 4-layered QRNN model with an embedding size of $400$ and $2500$ nodes in each hidden layer.
We trained the model using a batch size of $60$ and a BPTT length of $140$ using the Adam optimizer \cite{kingma2014adam} for $14$ epochs, reducing the learning rate by $10$ on epoch $12$.
To avoid over-fitting, we employ the regularization strategies proposed in \cite{merity2018regularizing} including variational dropout, random sequence lengths, and L2-norm decay.
The values for the model hyperparameters were tuned only coarsely; for full hyper-parameter values, refer to Table~\ref{table:fullhypers}.

\begin{table}[t]
\centering
\begin{tabular}{l|cc}
\toprule
\bf Model & \bf Val &  \bf Test \\
\midrule
\citet{Grave2016} & -- & 48.7 \\
\citet{dauphin2016language}, 1 GPU & -- & $44.9$ \\
\citet{dauphin2016language}, 4 GPUs & -- & $37.2$ \\
\midrule
4 layer QRNN ($h=2500$), 1 GPU & $32.0$ & $33.0$ \\
\bottomrule
\end{tabular}
\caption{Perplexity on the word-level WikiText-103 dataset.
The model was trained for 12 hours (14 epochs) on an NVIDIA Volta.
\label{table:wikitext103}
}
\end{table}

\begin{table}[t]
\centering
\begin{tabular}{l|r}
\toprule
\bf Model & \bf Time per batch \\
\midrule
LSTM & 726ms \\
QRNN & 233ms \\
\bottomrule
\end{tabular}
\caption{Mini-batch timings during training on WikiText-103.
This is a $3.1$ times speed-up over the NVIDIA cuDNN LSTM baseline which uses the same model hyperparameters.
\label{table:wt103time}}
\end{table}

\begin{table*}
\renewcommand{\arraystretch}{1.1}
\begin{center}
 \begin{tabular}{@{} lrrrrr @{}} 
 \toprule[1.5pt]
  &  \multicolumn{1}{c}{Character PTB} &  & \multicolumn{1}{c}{\enwik} & & WikiText-103 \\ 
 \midrule
 RNN Cell  & \small{LSTM} & & \small{LSTM} & & \small{QRNN}\\ 
 Layers & 3 & & 3 & & 4 \\
 RNN hidden size & 1000 & & 1840 & & 2500 \\
 AR/TAR & 0/0 & & 0/0 & & 0/0 \\
 Dropout (e/h/i/o) & 0/0.25/0.1/0.1 & & 0/0.01/0.01/0.4 & & 0/0.1/0.1/0.1 \\
 Weight drop & 0.5 & & 0.2 & & 0 \\
 \midrule
 Weight decay & $1.2\mathrm{e}{-6}$ & & $1.2\mathrm{e}{-6}$ & & 0 \\
 BPTT length & 150 & & 200 & & 140 \\
 Batch size & 128 & & 128 & & 60 \\
 Input embedding size & 128 & & 400 & & 400 \\
 Learning rate & 0.002 & & 0.001 & & 0.001 \\
 Epochs & 500 & & 50 & & 14 \\
 LR reduction ($\frac{lr}{10}$) & [300, 400] & & [25, 35] & & [12] \\
 \midrule
 Total parameters & 13.8M & & 47M & & 151M \\
 Training time (hours) & 8 & & 47 & & 12 \\
 \bottomrule[1.5pt]
\end{tabular}
\caption{
Hyper-parameters for word- and character-level language modeling experiments.
Training time is for all noted epochs on an NVIDIA Volta.
Dropout refers to embedding, (RNN) hidden, input, and output. \label{table:fullhypers}}
\end{center}
\end{table*}

\section{Analysis}

\subsection{QRNN vs LSTM}
QRNNs and LSTMs operate on sequential data in vastly different ways.
For word-level language modeling, QRNNs have allowed for similar training and generalization outcomes at a fraction of the LSTM's cost \citep{merity2018regularizing}.
In our work however, we have found QRNNs less successful at character-level tasks, even with substantial hyperparameter tuning.

\begin{figure}
\centering
\includegraphics[width=1\linewidth]{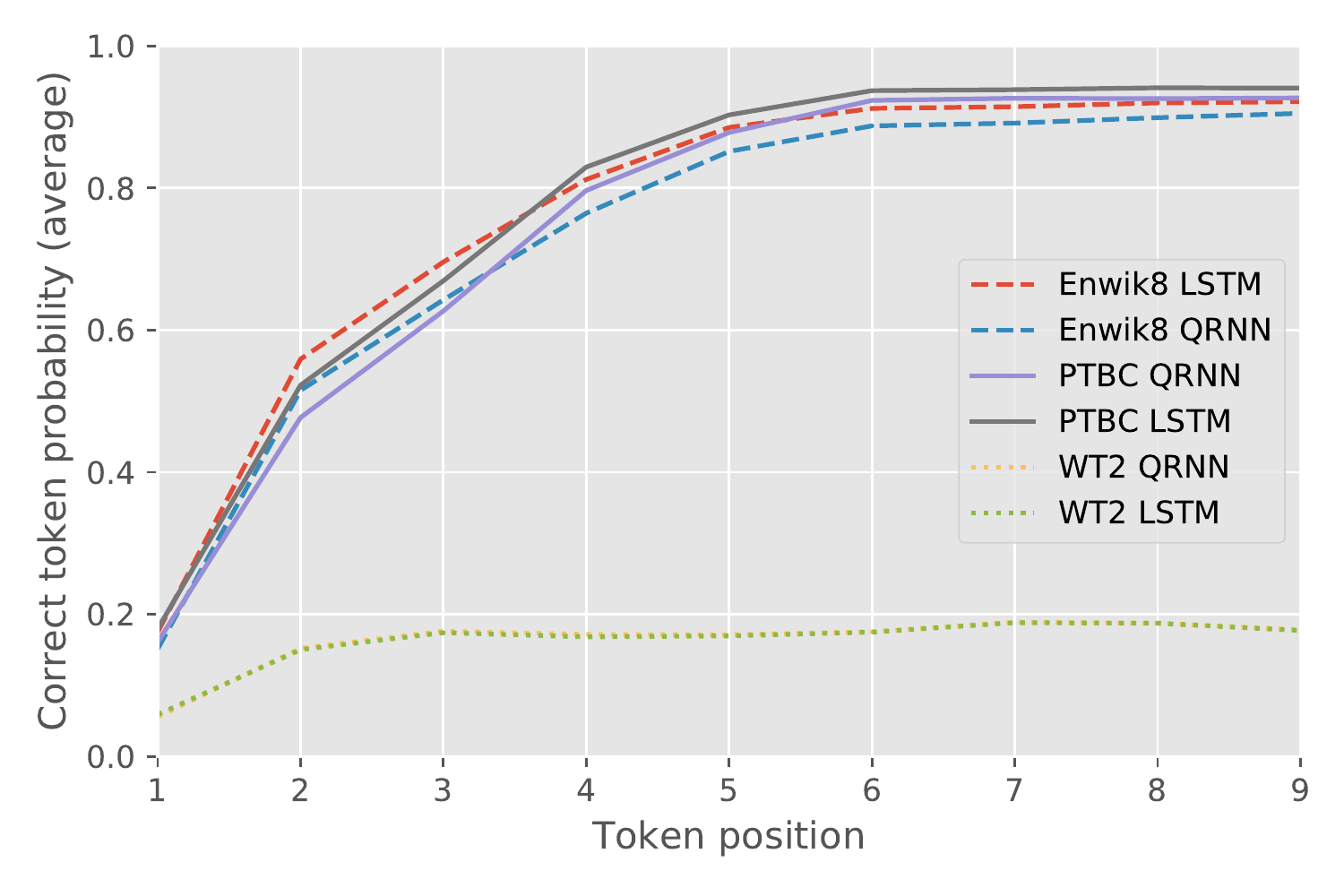}
\caption{
To understand how the models respond to uncertainty in different datasets, we visualize probability assigned to the correct token after a defined starting point. 
For character datasets (\enwik and PTB character-level), token position refers to characters within space delimited words.
For word level datasets, token position refers to words following the word \textit{The}, approximating the start of a sentence.
}
\label{fig:confusion}
\end{figure}

To investigate this, Figure \ref{fig:confusion} shows a comparison between both word- and character-level tasks as well as between the LSTM and the QRNN. In this experiment, we plot the probability assigned to the correct token as a function of the token position with the model conditioned on the ground-truth labels up to the token position. 
We attempt to find similar situations in the word- and character-level datasets to see how the LSTM and QRNN models respond differently.
For character-level datasets, model confusion is highest at the beginning of a word, as at that stage there is little to no information about it.
This confusion decreases as more characters from the word are seen.
The closest analogy to this on word-level datasets may be just after the start of a sentence.
As a proxy for finding sentences, we find the token \textit{The} and record the model confusion after that point.
Similarly to the character-level dataset, we assumed that confusion would be highest at this early stage before any information has been gathered, and decreased as more tokens are seen.
Surprisingly, we can clearly see the behavior of the datasets is quite different.
Word-level datasets do not gain the same clarity after seeing additional information compared to the character-level datasets.
We also see the QRNN underperforming the LSTM on both the Penn Treebank and \enwik character-level datasets.
This is not the case for the word-level datasets.

Our hypothesis is that character-level language modeling requires a more complex hidden-to-hidden transition.
As the LSTM has a full matrix multiplication between timesteps, it is able to more quickly adapt to changing situations.
The QRNN on the other hand suffers from a simpler hidden-to-hidden transition function that is only element-wise, preventing full communication between hidden units in the RNN.
This might also explain why QRNNs need to be deeper than the LSTM to achieve comparable results, such as the 6 layer QRNN in Table \ref{table:ptb}, as the QRNN can perform more complex interactions of the hidden state by sending it to the next QRNN layer.
This may suggest why state-of-the-art architectures for character- and word-level language modeling can be quite different and not as readily transferable.

\begin{figure}
\centering
\includegraphics[width=1\linewidth]{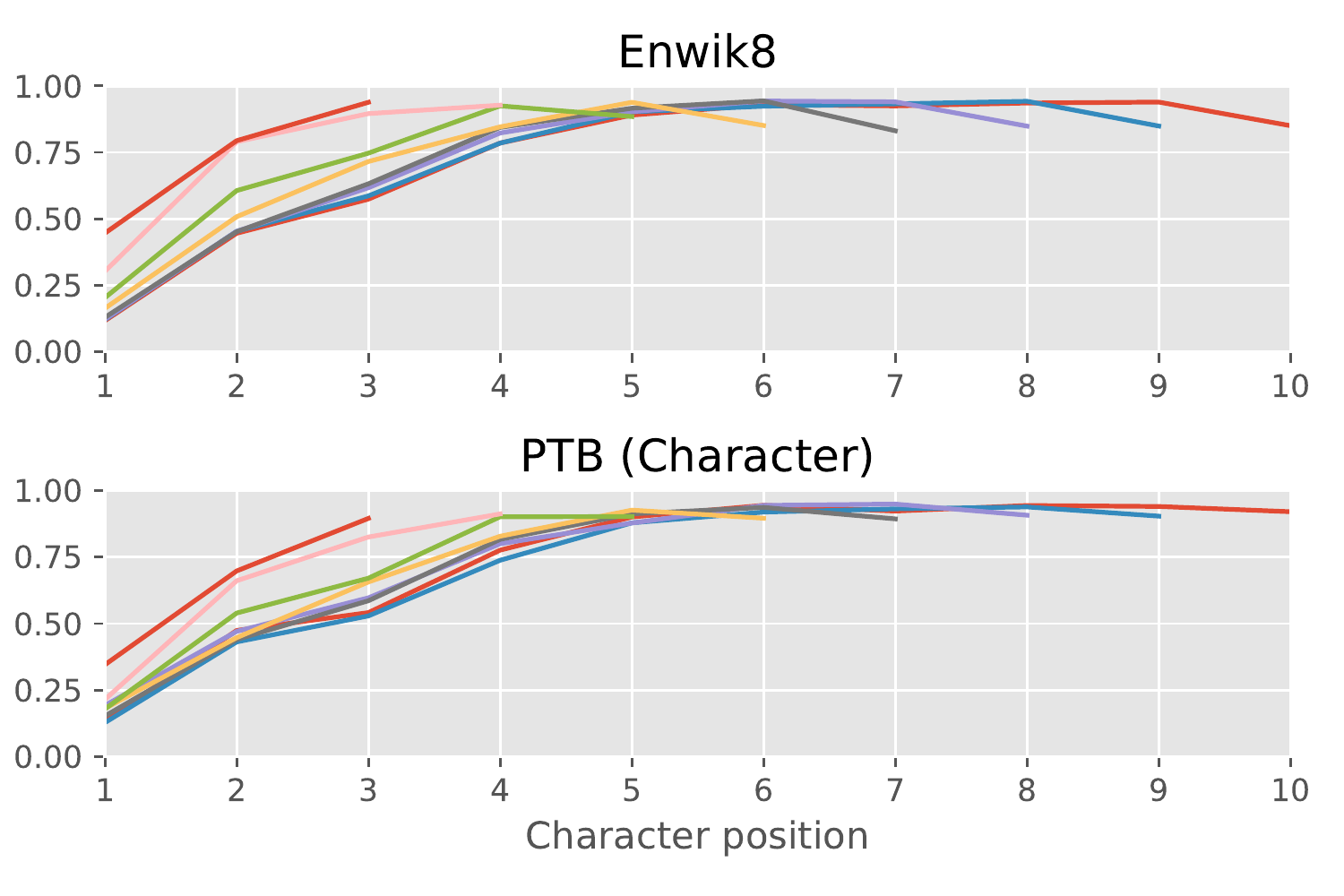}
\caption{
To better understand the character level datasets, we analyze the average probability of getting a character at different positions correct within a word.
Words are defined as two or more [A-Za-z] characters between spaces and different length words have different colors.
We include the prediction of the final space as it allows for capturing the uncertainty of when to end a word (i.e. \texttt{puzzle\_} vs \texttt{puzzles\_}).
Due to the bounded vocabulary of PTBC, we can see the model is far more confident in a word's ending than for model for \enwik.
}
\label{fig:wordlen_confusion}
\end{figure}

\subsection{Hyperparameter importance}
\begin{figure}
\centering
\includegraphics[width=1\linewidth]{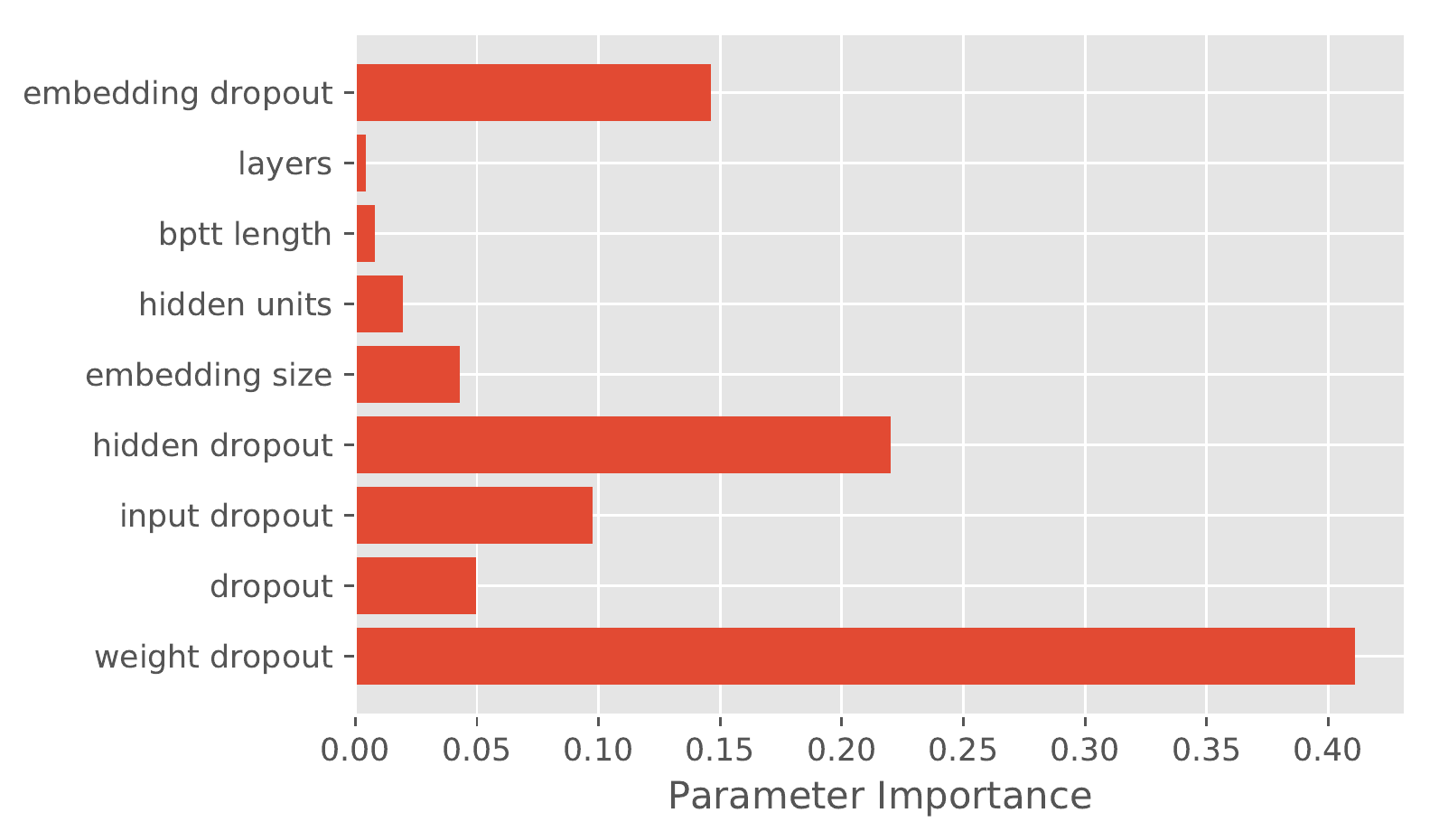}
\caption{
We analyze the relative importance of the hyperparameters defining the model using a Random Forest approach for the word-level task on the smaller WikiText-2 data set for AWD-QRNN model. The results show that weight dropout, hidden dropout and embedding dropout impact performance the most while the number of layers and the embedding and hidden dimension sizes matters relatively less. Similar results are observed on the PTB word level data set. 
}
\label{fig:hyperimp}
\end{figure}

Given the large number of hyperparameters in most neural language models, the process of tuning models for new datasets can be laborious and expensive. In this section, we attempt to determine the relative importance of the various hyperparameters. To do this, we train $200$ models with random hyperparameters and then regress them against the validation perplexity using MSE-based Random Forests \cite{breiman2001random}. We then use the Random Forest's feature importance as a proxy for the hyperparameter importance. While the hyperparameters are random, we place natural bounds on them to prevent unrealistic models given the AWD-QRNN framework and its recommended hyperparameters. For this purpose, all dropout values are bounded to $[0,1]$, the truncated BPTT length is bounded between $30$ and $300$, number of layers are bounded between $1$ and $10$ and the embedding and hidden sizes are bounded between $100$ and $500$. We present the results for the word-level task on the WikiText-2 data set using the AWD-QRNN model in Figure~\ref{fig:hyperimp}. The models were trained for 300 epochs with Adam with an initial learning rate of $10^{-3}$ which is reduced by $10$ on epochs $150$ and $225$. The results show that weight dropout, hidden dropout and embedding dropout impact performance the most while the number of layers and the sizes of the embedding and hidden layers matter relatively less. Experiments on the Penn Treebank data set also yielded similar results. Given these results, it is evident that in the presence of limited tuning resources, educated choices can be made for the layer sizes and the dropout values can be finely tuned. 

\begin{figure*}
    \centering
    \begin{subfigure}[b]{0.3\linewidth}
        \includegraphics[width=\linewidth]{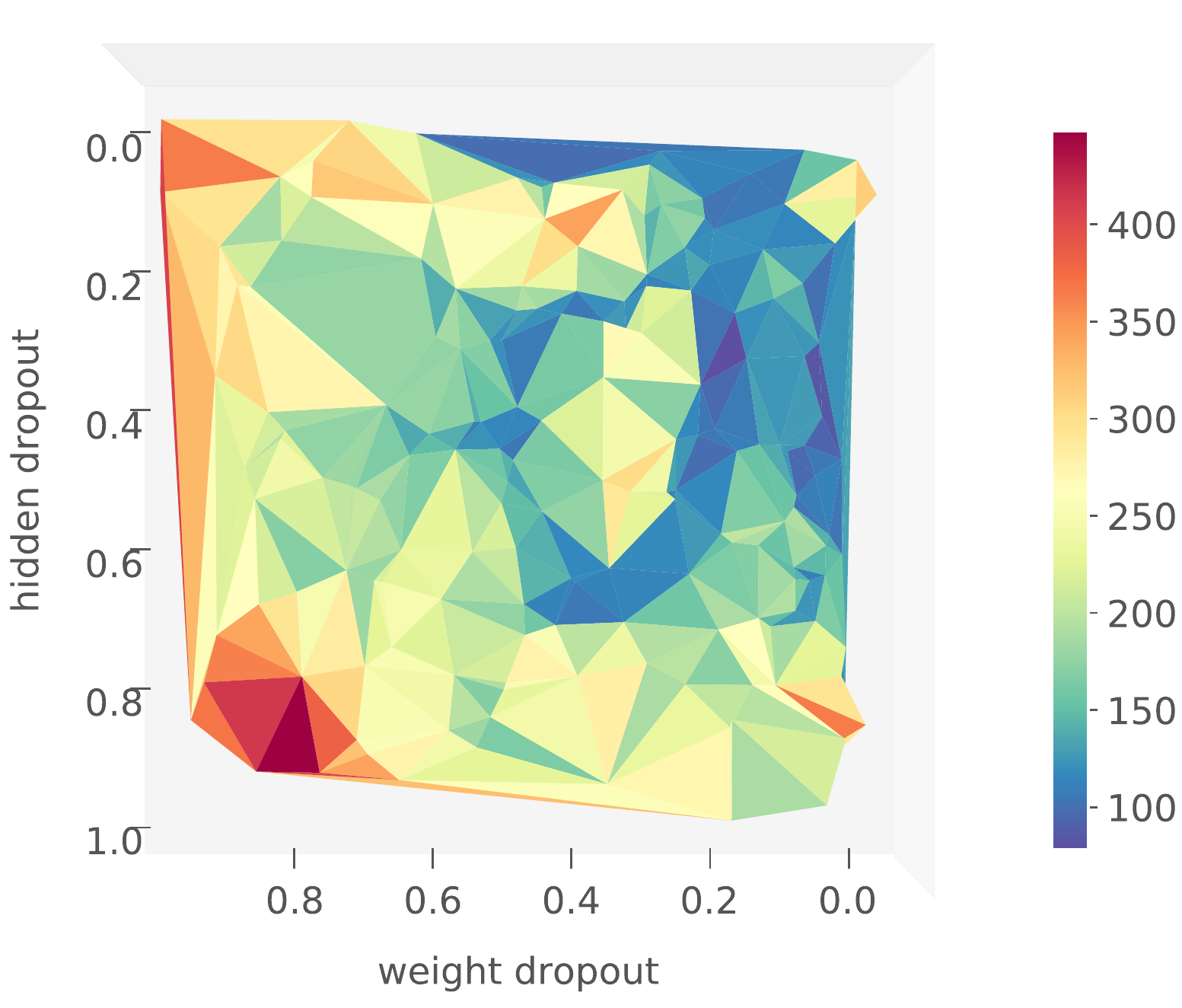}
        \centering
        \caption{Joint influence of weight dropout and hidden-to-hidden dropout.}
        \label{fig:gull}
    \end{subfigure}
    \qquad
\begin{subfigure}[b]{0.3\linewidth}
        \includegraphics[width=\linewidth]{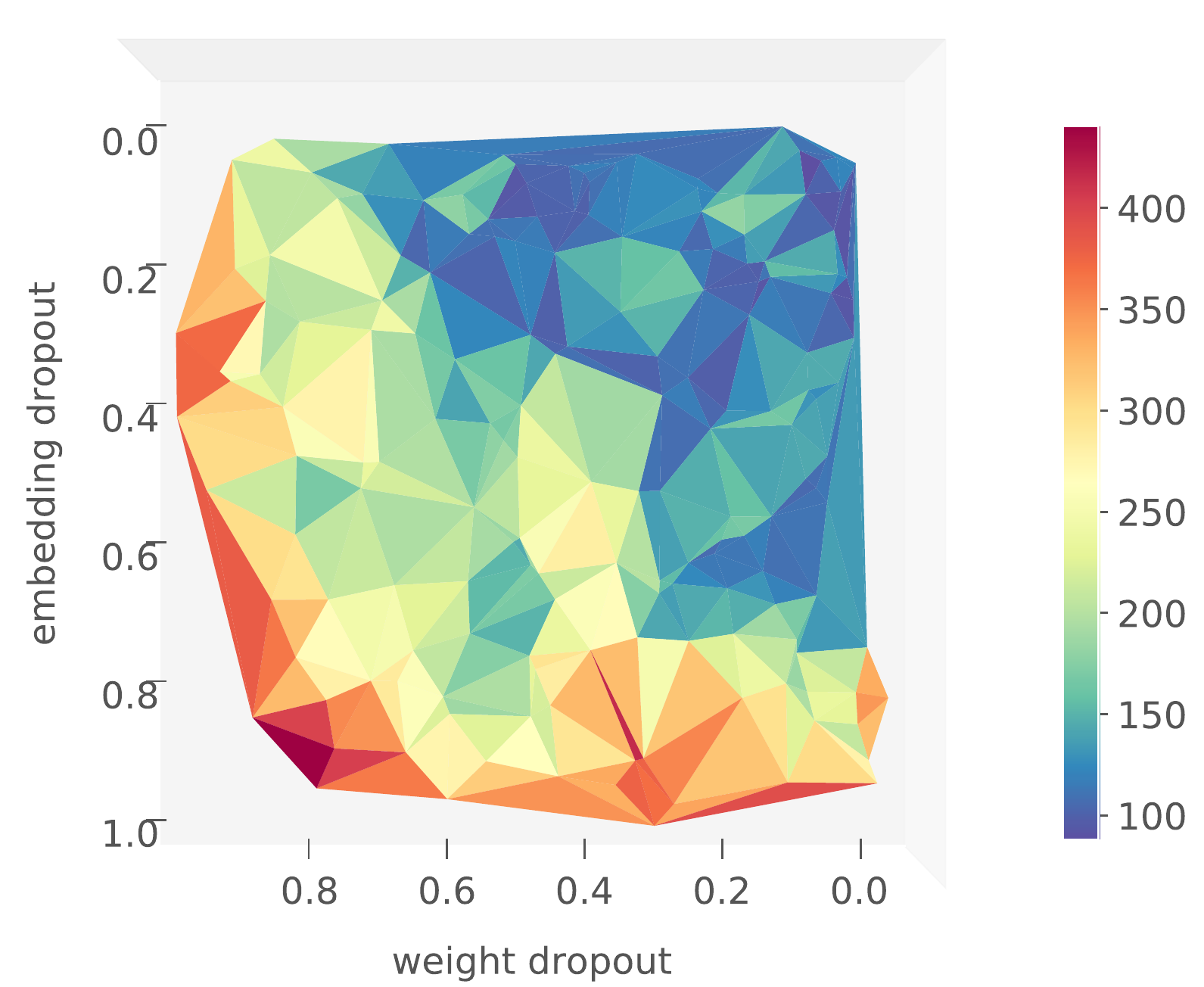}
        \centering
        \caption{Joint influence of weight dropout and embedding dropout.}        
        \label{fig:tiger}
    \end{subfigure}
    \qquad
\begin{subfigure}[b]{0.3\linewidth}
        \includegraphics[width=\linewidth]{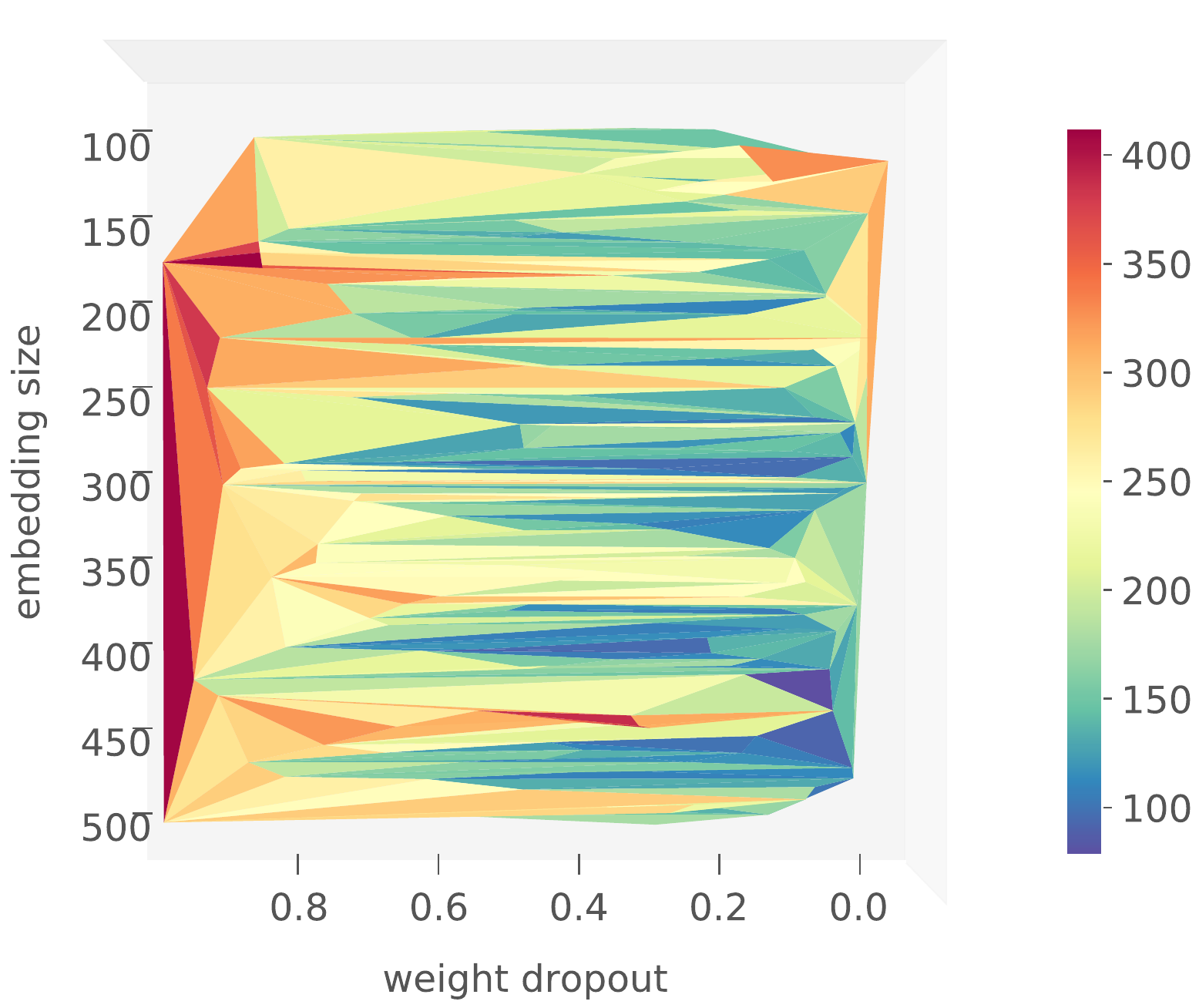}
        \centering
        \caption{Joint influence of weight dropout and embedding size.}        
        \label{fig:tiger}
    \end{subfigure}
    
    \caption{Heatmaps plotting the joint influence of three sets of hyperparameters (weight dropout - hidden dropout, weight dropout - embedding dropout and  weight dropout - embedding size) on the final perplexity of the AWD-QRNN language model trained on the WikiText-2 data set. Results show ranges of permissible values for both experiments and the strong coupling between them. For the hidden dropout experiment, results suggest a narrower band of acceptable values while the embedding dropout experiment suggests a more forgiving dependence so long as the embedding dropout is kept low. The joint plot between weight dropout (high influence) and embedding size (low influence) suggests relative insensitivity to the embedding size so long as the size is not too small or too large.\label{fig:slice}}
\end{figure*}

In Figure~\ref{fig:slice}, we plot the joint influence heatmaps of pairs of parameters to understand their couplings and bounds. For this experiment, we consider three pairs of hyperparameters: weight dropout - hidden dropout, weight dropout - embedding dropout and  weight dropout - embedding size and plot the perplexity, obtained from the the WikiText-2 experiment above, in the form of a projected triangulated surface plot. The results suggest a strong coupling for the high-influence hyperparameters with narrower bounds for acceptable performance for hidden dropout as compared to weight dropout. The low influence of the embedding size hyperparameter is also evident from the heatmap; so long as the embeddings are not too small or too large, the influence of this hyperparameter on the performance is not as drastic. Finally, the plots also suggest that an educated guess for the dropout values lies in the range of $[0.1,0.5]$ and tuning the weight dropout first (to say, $0.2$) leaving the rest of the hyperparameters to estimates would be a good starting point for fine-tuning the model performance further. 

\section{Discussion}

\subsection{Penn Treebank is flawed for character-level work}

While the Mikolov processed Penn Treebank data set has long been a central dataset for experimenting with language modeling, we poset that it is fundamentally flawed.
As noted earlier, the character-level dataset does not feature punctuation, capitalization, or numbers - all important aspects we would wish for a language model to capture. 
The limited vocabulary and use of \verb+<unk>+ tokens also result in substantial problems.

In Figure \ref{fig:wordlen_confusion} we compare the level of average surprise between two models when producing words of a given length: one trained on Penn Treebank and the other trained on \enwik.
\enwik has a noticeable drop when predicting the final character (the space) compared to the Penn Treebank results.
The only character transitions that exist within the Penn Treebank dataset are those contained within the 10,000 word vocabulary.
Models trained on the \enwik dataset can't have equivalent confidence as words can branch unpredictably due to an unrestricted vocabulary.
In addition, the lack of capitalization, punctuation, or numerals removes many aspects of language that should be fundamental to the task of character-level language modeling.
Potentially counter-intuitively, we do not recommend its use as a benchmark even though we achieve state-of-the-art results on this dataset.

\subsection{Parameter count as a proxy for complexity}

To measure the complexity and computational cost of training and evaluating models, the number of parameters in a model are often reported and compared.
For certain use cases, such as when a given model is intended for embedded hardware, parameter counts may be an appropriate metric.
Conversely, if the parameter count is unusually high and may require substantial resources \citep{Shazeer2017}, the parameter count may still function as an actionable metric.

In general however a model's parameter count acts as a poor proxy for the complexity and hardware requirements of the given model.
If a model with high parameter count runs quickly and on modest hardware, we would argue this is better than a model with a lower parameter count that runs slowly or requires more resources.
More generally, parameter counts may be discouraging proper use of modern hardware, especially when the parameter counts were historically motivated by a now defunct hardware requirement.

\section{Related work}

Our work is not the first work to show that well tuned and fast baseline models can be highly competitive with state-of-the-art work.

In \citet{Melis2017}, several state-of-the-art language model architectures are re-evaluated using large-scale automatic black-box hyper-parameter tuning.
Their results show that a standard LSTM baseline, when properly regularized and tuned, can outperform many of the recently proposed state-of-the-art models on a word-level language modeling tasks (PTB and WikiText-2).

\citet{Merity2016} proposes the weight-dropped LSTM, which uses DropConnect on hidden-to-hidden weights as a form of recurrent regularization, and NT-ASGD, a variant of the averaged stochastic gradient method.
Applying these two techniques to a standard LSTM language modeling baseline, they achieve state-of-the-art results similarly to \citet{Melis2017}.

The most recent results on character-level datasets generally involve more complex architectures however.
\citet{fastslowlm} introduce the Fast-Slow RNN, which splits the standard language model architecture into a fast changing RNN cell and a slow changing RNN cell.
They show the slow RNN's hidden state experiences less change than that of the fast RNN, allowing for longer term dependency similar to that of multiscale RNNs.

The Recurrent Highway Network \citep{Zilly2016} focuses on a deeper hidden-to-hidden transition function, allowing the RNN to spend more than one timestep processing a single input token.

Additional improvements in the model can be obtained through dynamic evaluation \cite{dynamiceval} and mixture-of-softmaxes \cite{softmaxmixture} but since our goal is to evaluate the underlying model, we employ no such strategies in addition.

\section{Conclusion}

Fast and well tuned baselines are an important part of our research community.
Without such baselines, we lose our ability to accurately measure our progress over time.
By extending an existing state-of-the-art word level language model based on LSTMs and QRNNs, we show that a well tuned baseline can achieve state-of-the-art results on both character-level (Penn Treebank, \enwik) and word-level (WikiText-103) datasets without relying on complex or specialized architectures.
We additionally perform an empirical investigation of the learning and network dynamics of both LSTM and QRNN cells across different language modeling tasks, highlighting the differences between the learned character and word level models. Finally, we present results which shed light on the relative importance of the various hyperparameters in neural language models. On the WikiText-2 data set, the AWD-QRNN model exhibited higher sensitivity to the hidden-to-hidden weight dropout and input dropout terms and relative insensitivity to the embedding and hidden layer sizes. We hope that this insight would be useful for practitioners intending to tune similar models on new datasets. 

\newpage
\bibliography{example_paper}
\bibliographystyle{icml2018}

\end{document}